\documentclass[11pt,a4paper]{article}
\usepackage[hyperref]{acl2021}
\usepackage{times}
\usepackage{latexsym}
\usepackage{tabu}
\usepackage{xingstyle}
\usepackage{graphicx}
\usepackage{amsmath}
\usepackage{mathtools}
\usepackage{mathrsfs}
\usepackage{hhline}
\usepackage{textcomp}
\usepackage{multirow}

\newcommand{\D}{\mathcal{D}}

\renewcommand{\L}{\mathcal{L}}

\aclfinalcopy 
\def\aclpaperid{17} 

\title{Multi-Pair Text Style Transfer on Unbalanced Data}

\author{Xing Han \\
University of Texas at Austin \\
Austin, TX, 78712 \\
\texttt{aaronhan223@utexas.edu} \\
\And 
Jessica Lundin \\
Salesforce \\
San Francisco, CA, 94105 \\
\texttt{jlundin@salesforce.com} \\}

\date{}

\begin{document}
\maketitle
\begin{abstract}
Text-style transfer aims to convert text given in one domain into another by paraphrasing the sentence or substituting the keywords without altering the content. By necessity, state-of-the-art methods have evolved to accommodate nonparallel training data, as it is frequently the case there are multiple data sources of unequal size, with a mixture of labeled and unlabeled sentences. Moreover, the inherent style defined within each source might be distinct. A generic bidirectional (e.g., formal $\Leftrightarrow$ informal) style transfer regardless of different groups may not generalize well to different applications. In this work, we developed a task adaptive meta-learning framework that can simultaneously perform a multi-pair text-style transfer using a single model. The proposed method can adaptively balance the difference of meta-knowledge across multiple tasks. Results show that our method leads to better quantitative performance as well as coherent style variations. Common challenges of unbalanced data and mismatched domains are handled well by this method.
\end{abstract}
\section{Introduction}
\begin{figure*}[t!]
\begin{minipage}{\textwidth}
    \centering
        \begin{tabular}{c @{\hspace{-1ex}} c @{\hspace{-1ex}} c}
        \begin{tabular}{c}
        \includegraphics[width=.31\textwidth]{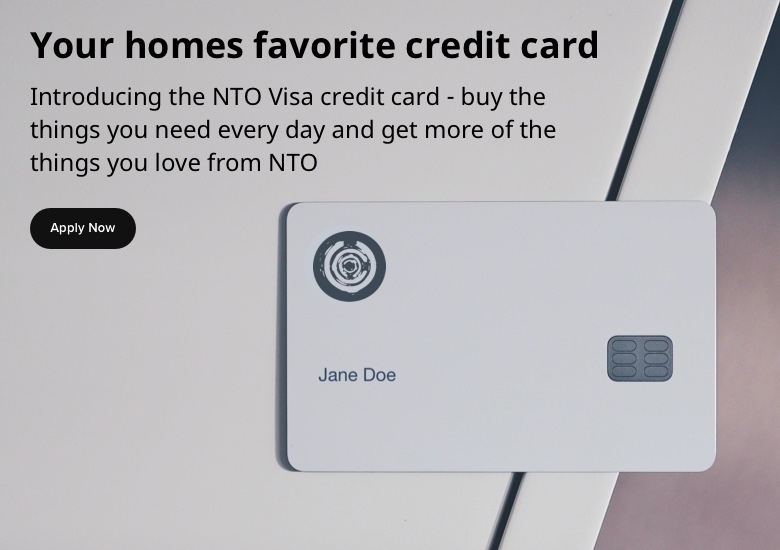}
        \\
        {\small{(a) Credit Card - Original}}
        \end{tabular} &
        \begin{tabular}{c}
        \includegraphics[width=.31\textwidth]{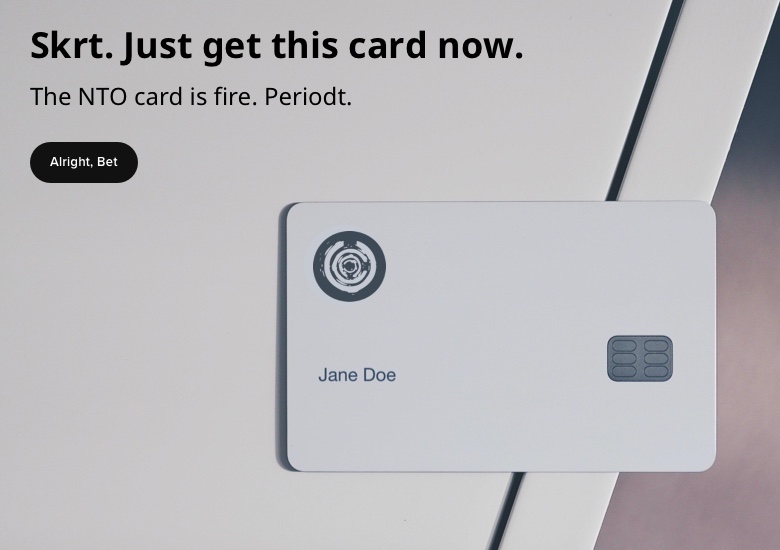}
        \\
        {\small{(b) Credit Card - Informal}}
        \end{tabular} & 
        \begin{tabular}{c}
        \includegraphics[width=.31\textwidth]{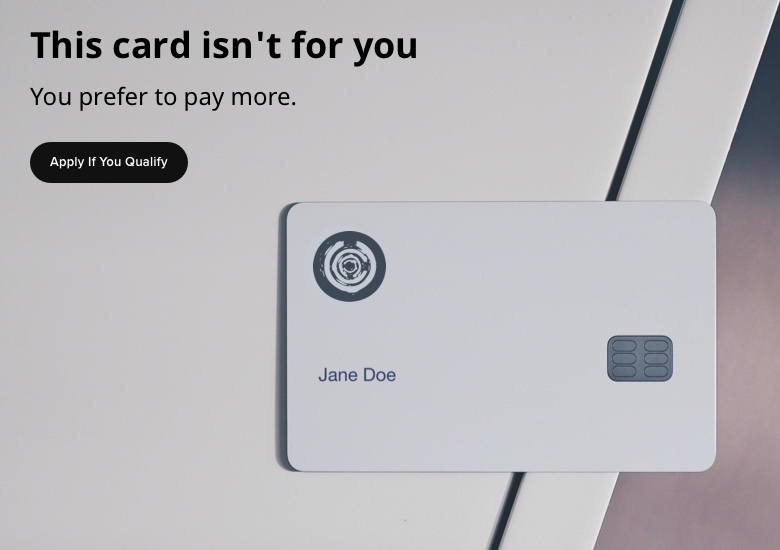} 
        \\
        {\small{(c) Credit Card - Exclusive}}
        \end{tabular} \\
        \begin{tabular}{c}
        \includegraphics[width=.31\textwidth]{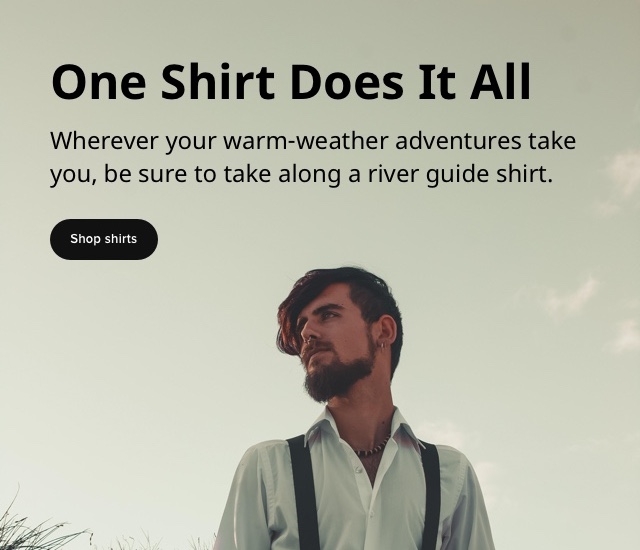}
        \\
        {\small{(d) Shirt - Original}}
        \end{tabular} &
        \begin{tabular}{c}
        \includegraphics[width=.31\textwidth]{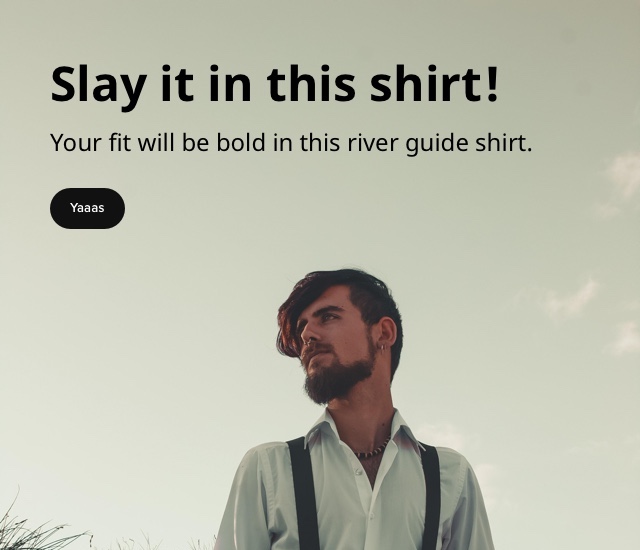}
        \\
        {\small{(e) Shirt - Informal}}
        \end{tabular} & 
        \begin{tabular}{c}
        \includegraphics[width=.31\textwidth]{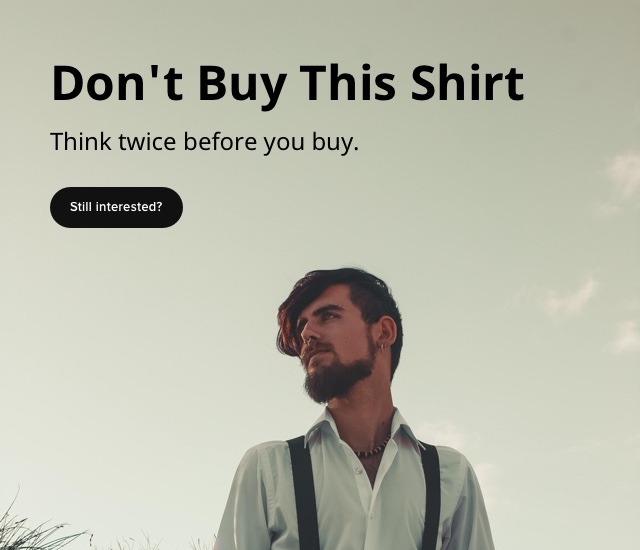} 
        \\
        {\small{(f) Shirt - Exclusive}}
        \end{tabular} \\
        \end{tabular}
\end{minipage}
\caption{Text style transfer examples in generative design: the original text is meta-data from e-commerce websites. Two target style variations are predefined for certain groups of customers.}
\label{fig:example}
\end{figure*}

Text-style transfer is a fundamental challenge in natural language processing. Applications include non-native speaker assistants, child education, personalization and generative design \citep{fu2017style, zhou2017mechanism, yang2018stylistic, gatys2016image, gatys2016preserving, zhu2017unpaired, li2017demystifying}. Figure \ref{fig:example} shows a prominent example on applying style transfer into a hypothetical online shopping platform, where the generated style variations can be used for personalized recommendations. However, compared with other domains, the lack of parallel corpus and quality training data is currently an obstacle for text-style transfer research. For example, assume one supports a multi-tenant service platform including tenant-specific text data, but there is no guarantee that each tenant will provide sufficient amount of data for model training. To build a multi-task language model that matches the text-style of each tenant is more practical and efficient than training individual models. This single-model approach might also have relatively favorable empirical performance.  


Existing works on text style transfer have addressed different applications such as sentiment transfer \citep{shen2017style}, word decipherment \citep{knight2006unsupervised}, and author imitation \citep{xu2012paraphrasing}. If parallel training data is available, a wide range of supervised techniques in machine translation (e.g., Seq2Seq models \citep{bahdanau2014neural} and Transformers \citep{vaswani2017attention}) can also be applied to style transfer problems. For non-parallel data, \citet{he2020probabilistic} proposed a probabilistic formulation that models non-parallel data from two domains as a partially observed parallel corpus, and learn the style transfer model in a completely unsupervised fashion. Unsupervised machine translation method has also been adapted to this setting \citep{zhang2018style}. In recent research focused on learning disentangled content and style representations using adversarial training \citep{john2018disentangled, yang2018unsupervised, shen2017style}, models are designed for non-parallel data while preserving content. \citet{lample2018multiple} argued that the adversarial models are not really doing disentanglement, and proposed a denoising auto-encoding approach instead. Another way to approach this problem is through identifying and substituting style-related sub-sentences \citep{li2018delete, sudhakar2019transforming}, where the unchanged part guarantees consistency over content. Additionally, state-of-the-art language models (BERT \citep{devlin2018bert}, GPT-2 \citep{radford2019language}, CTRL \citep{keskar2019ctrl}, etc.) and text-to-text models \citep{raffel2019exploring} achieve good performance generating text in different styles on multiple tasks \citep{dathathri2019plug, Wolf2019HuggingFacesTS}.

Building upon previous work, we aim to bridge real applications while accounting for the aforementioned data problems. Specifically, we wish to design an efficient training method for a style transfer model that 1. quickly learns and adapts to different style domains with limited data; 2. handling class-imbalance and out-of-distribution tasks. To achieve this, we introduce meta-learning into the style-transfer problem. 

\begin{figure*}[t!]
    \centering
    \includegraphics[width=\linewidth]{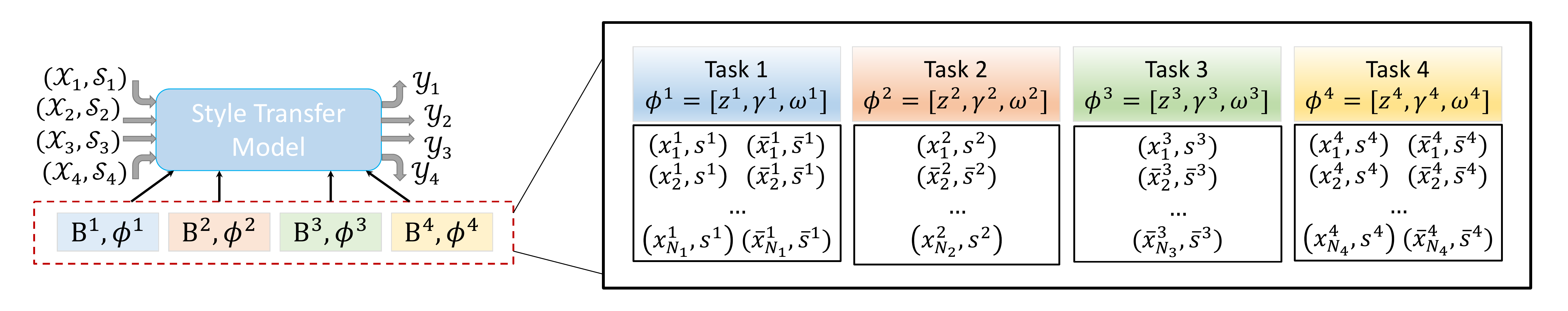}
    \caption{An overview of our multi-pair style transfer method: assume learning from each tenant's data is a task, and the training data available for each task varies. The style transfer model can adaptively learn tasks using our method and the resulting model performs style transfer across multiple domains.}
    \label{fig:showcase}
\end{figure*}

Meta-learning \citep{schmidhuber1987evolutionary} is a method to enable generalization ability to a model over a distribution of tasks. We focus on optimization-based meta-learning for our applications. MAML \citep{finn2017model} learns a common initialization parameter for each task using a few gradient steps. This standard MAML approach has been applied to text style transfer problems with low resources \citep{chen2020st} and achieved better performance in this situation. However, this method did not take into account the internal variations between tasks. A similar algorithm called Reptile \citep{nichol2018first} achieves better performance by maximizing the inner product between gradient of different mini-batches from the same task in its update. Recent works \citep{qiao2018few, lee2018gradient} improved a single meta-learner to task-adaptive meta-learning models, which includes task-specific parameters to help generalize better between tasks. Bayesian meta-learning is another active area of research: \citet{finn2018probabilistic} proposed a probabilistic version of MAML, where the variational inference framework utilizes a task-specific gradient update. More recently, \citet{lee2019learning} incorporated a Bayesian framework into task-adaptive meta-learning. Specifically, they introduce balancing variables for task and class-specific learning and leverage the uncertainties of these parameters derived from training data statistics. In this paper, we will adapt the Bayesian task adaptive meta-learning (TAML) for our application shown in Figure \ref{fig:showcase} overview. 
\section{Balancing Variations between Tasks}

A common challenge in aforementioned real application is that data from multiple sources may suffer from different problems, such as insufficient training samples, unbalanced class labels, or domain mismatch. However, simply ignore these differences and concatenate all tenants' data for model training will not lead to ideal results. 

Meta-learning is one of the most relevant approaches for generalized learning from few samples of different tasks. Assume a task distribution $p(\tau)$ that randomly generates task $\tau$ consisting of training set $\D^{\tau} = \{X^{\tau}, \overline{X}^{\tau}\}$ and a test set $\D^{\tau}_{\mathsf{test}} = \{X^{\tau}_{\mathsf{test}}, \overline{X}^{\tau}_{\mathsf{test}}\}$. If parallel training data is not available, then we only have $\D^{\tau} = X^{\tau}$ and $\D^{\tau}_{\mathsf{test}} = X^{\tau}_{\mathsf{test}}$. The MAML algorithm initialize task-specific parameter $\theta^{\tau}$ using a few gradient steps on a small amount of data. In this case, the optimized parameters can generalize to new tasks. Specifically, we have the loss minimization
\begin{equation}
	\underset{\theta}{\min} \underset{\tau \sim p(\tau)}{\sum} \L(\theta - \alpha \nabla_{\theta} \L(\theta; \D^{\tau}); \D^{\tau}_{\mathsf{test}}), 
	\label{eq:maml}
\end{equation}
where $\alpha$ is the step size when learning each task. The initial parameter of each task then becomes $\theta^{\tau} = \theta - \alpha \nabla_{\theta} \L(\theta; \D^{\tau}),$ which has been proved to minimize the test loss $\L(\theta^{\tau} ; \D^{\tau}_{\mathsf{test}})$. The training set $\D^{\tau}$ may consist of only a few samples.

Eq (\ref{eq:maml}) is effective in numerous applications, yet insufficient in addressing our data problems, as it treats the initialization and learning parameters with equal importance for each task. Inspired by \citet{lee2019learning}, we now introduce three balancing variables: $z^{\tau}, \gamma^{\tau}, \omega^{\tau}$ for every task $\tau$. 

Let ${\omega}^{\tau} = (\omega_1^{\tau}, ..., \omega_C^{\tau}) \in [0, 1]^C$ be the multiplier of each of the class specific gradients to vary the learning rate for each class. In real applications, we often have a style transfer problem with unbalanced training data. For instance, when training formality style transfer models, the number of formal/positive sentences is normally much larger than the number of informal/exclusive sentences. Also, denote $\gamma^{\tau} = (\gamma_1^{\tau}, ..., \gamma_L^{\tau}) \in [0, \infty)^L$ to be the multipliers of the original learning-rate $\alpha$, where the new learning rate becomes $\gamma_1^{\tau} \alpha, \gamma_2^{\tau} \alpha, ..., \gamma_L^{\tau} \alpha$. Note that the value of $\gamma$ is task-dependent (e.g., sample size of the training data from each task), and is meant to deal with the small data problem in multi-pair text style transfer. Moreover, since the text data collected from every source or tenant is very hard to be aligned, it is common to have training data with significantly different context. We can treat this as an out of distribution problem and this can be reflected on the value of initial parameters. We use $z^{\tau}$ to modulate the initial parameter $\theta$ for each task. Specifically, $z^{\tau}$ relocates the initial $\theta$ to a task-dependent starting point prior to the learning process. We unify these properties as the learning framework below:

\begin{align}
	\theta_0 & = \theta * z^{\tau}, \quad \mathsf{and ~ for} ~ k = 1,..., \mathsf{K}: \label{eq:taml} \\
	\theta_k & = \theta_{k-1} - \gamma^{\tau} \circ \alpha \circ \sum_{c=1}^C \omega_c^{\tau} \nabla_{\theta_{k-1}} \L(\theta_{k-1}; \D_c^{\tau}), \nonumber
\end{align}
where $\omega_c$ and $\D_c$ are class-specific parameters and data; $\mathsf{K}$ is the total number of iterations for updating parameters. We currently assume $C=2$ in the following discussions of this paper, since pair-wise style transfer is the primary problem of interest so far.

\begin{figure*}[t!]
    \centering
    \includegraphics[width=\linewidth]{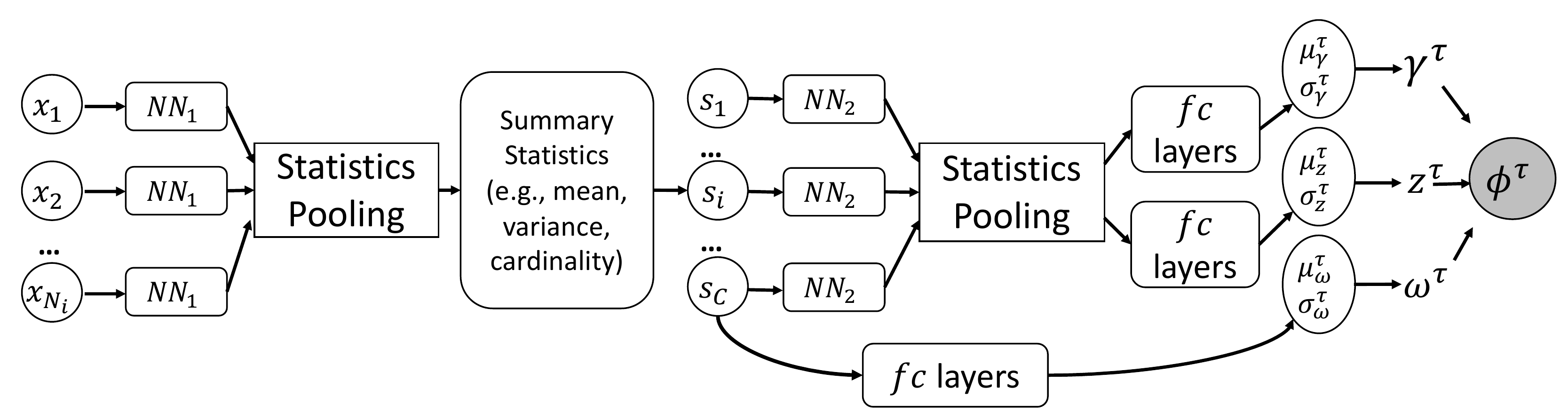}
    \caption{The inference network for generating posterior distribution of balancing variables $\gamma, z,$ and $\omega$ for task $i$.}
    \label{fig:inference}
\end{figure*}

\section{Learning the Balancing Variables through Variational Inference}
We now discuss how to find the most suitable value of each balancing variable. We employ the variational inference framework from probabilistic MAML \citep{he2020probabilistic} and TAML \citep{lee2019learning} to extract the task-specific information. The variational inference framework is used to compute posterior distributions for the balancing variables $z^{\tau}, \gamma^{\tau}, \omega^{\tau}$. Assume the training data $X^{\tau} = \{x_n^{\tau}\}_{n=1}^{N_{\tau}}$, $\overline{X}^{\tau} = \{\overline{x}_n^{\tau}\}_{n=1}^{N_{\tau}}$; test data $X^{\tau}_{\mathsf{test}} = \{x_m^{\tau}\}_{m=1}^{M_{\tau}}$, $\overline{X}^{\tau}_{\mathsf{test}} = \{\overline{x}_m^{\tau}\}_{m=1}^{M_{\tau}}$, and $\phi^{\tau} = \{\tilde{\omega}^{\tau}, \tilde{\gamma}^{\tau}, \tilde{z}^{\tau}\}$ to be a collection of three balancing variables. The goal of learning for each task $\tau$ is to maximize the conditional log-likelihood of the joint dataset $\D^{\tau}_{\mathsf{test}}$ and $\D^{\tau}$: $\log p(\overline{X}^{\tau}_{\mathsf{test}}, \overline{X}^{\tau} | X^{\tau}_{\mathsf{test}}, X^{\tau}; \theta)$. To solve the optimization problem requires determining the true posterior $p(\phi^{\tau}|\D^{\tau}, \D^{\tau}_{\mathsf{test}})$, which is intractable. We resort to variational inference with a tractable form of approximate posterior $q(\phi^{\tau}|\D^{\tau}, \D^{\tau}_{\mathsf{test}}; \psi)$ parameterized by $\psi$. In order to make the inference network of meta training and meta testing consistent, we drop the dependency of $\D^{\tau}_{\mathsf{test}}$ since the test labels are unknown in meta-testing. Hence the approximate posterior becomes $q(\phi^{\tau}|\D^{\tau}; \psi)$. We now have the approximated lower bound for task adaptive meta learning:
\begin{align}
	&\L_{\theta, \psi}^{\tau} = \frac{N_{\tau} + M_{\tau}}{M_{\tau}} \sum_{m=1}^{M_{\tau}} E_{q(\phi^{\tau}|\D^{\tau}; \psi)} \label{eq:obj} \\
	&\left[\log p(\tilde{y}_m^{\tau} | \tilde{x}_m^{\tau}, \phi^{\tau}; \theta)\right] - \mathsf{KL}[q(\phi^{\tau}|\D^{\tau}; \psi) \| p(\phi^{\tau})].
	\nonumber
\end{align}
Given that each balancing variable is independent, $q(\phi^{\tau}|\D^{\tau}; \psi)$ can therefore be fully factorized
\begin{align*}
	&q(\phi^{\tau}|\D^{\tau}; \psi) = \\
	&\prod_c q(\omega_c^{\tau} | \D^{\tau}; \psi) \prod_l q(\gamma_l^{\tau} | \D^{\tau}; \psi) \prod_i q(z_i^{\tau} | \D^{\tau}; \psi).
\end{align*}

We assume each single dimension of $q(\phi^{\tau} | \D^{\tau}; \psi)$ follows a uni-variate Gaussian distribution with trainable mean and variance. Given $\phi_s^{\tau} \sim q(\phi^{\tau} | \D^{\tau}; \psi)$, we then use the Monte-Carlo approximation on Eq (\ref{eq:obj}) as a new objective:
\begin{align} 
	& \underset{\theta, \psi}{\min} \frac{1}{M_{\tau}} \sum_{m=1}^{M_{\tau}} \frac{1}{S} \sum_{s=1}^S -\log p(\tilde{y}_m^{\tau} | \tilde{x}_m^{\tau}, \phi_s^{\tau}; \theta) \nonumber \\
	& + \frac{1}{N_{\tau} + M_{\tau}} \mathsf{KL}[q(\phi^{\tau}|\D^{\tau}; \psi) \| p(\phi^{\tau})].
	\label{eq:mc_approx}
\end{align}

To better model the variational distribution $q(\phi^{\tau}|\D^{\tau}; \psi)$, an informative representation encoded from the training dataset $\D^{\tau}$ is necessary. In this case, the inference network can capture all useful statistical information in $\D^{\tau}$ to recognize its imbalances. We use a two-stage hierarchical set encoder, for a given text style transfer task, we first encodes each class, and then encodes the whole set of classes. Define the encoder $\mathsf{StatisticsPooling(\cdot)}$ that generates concatenation of the class statistics such as mean, variance and cardinality. The two-stage encoder first encodes all text sentences of each class into $s_c$, followed by encoding representations of the whole set of classes:
\begin{align*}
	v^{\tau} &= \mathsf{StatisticsPooling\left(\{NN_2(s_c)\}_{c=1}^C\right)}, \\
	s_c &= \mathsf{StatisticsPooling\left(\{NN_1(x)\}_{x\in X_c^{\tau}}\right)},
\end{align*}
where $c = 1, ..., C$ represents classes; $X_c^{\tau}$ is the collection of class $c$ examples in task $\tau$; $\mathsf{NN_1}$ and $\mathsf{NN_2}$ are some neural networks parameterized by $\psi$. Therefore, the summarized feature vectors of $\D^{\tau}$ can be used to infer the Gaussian distribution parameters of balancing variables $\omega^{\tau}, z^{\tau}$ and $\gamma^{\tau}$ to be further applied in the update of meta-learning. Note that since the balancing variable $\omega$ is class-specific, inference its distributional parameters does not need to go through the second stage of encoding. The overall structure of the inference network is shown in Figure \ref{fig:inference}.


\begin{algorithm*}[t]
\small
\caption{Multi-Pair Text Style Transfer via TAML}
\begin{algorithmic}[1]
\State \textbf{Input:} style pair for each task $\tau$, $\{(s^{\tau}, \overline{s}^{\tau})\}_{\tau=1}^{\mathcal{T}}$, parameters $\alpha, \beta$,
\State \textbf{Meta-training procedure:}
\While{not done}
\For{each style pair $(s^{\tau}, \overline{s}^{\tau})$}
\State Train inference network $q(\phi^{\tau}|\D^{\tau}; \psi)$ by minimizing objective (\ref{eq:mc_approx})
\State Obtain balancing variables $\{z^{\tau}, \gamma^{\tau}, \omega^{\tau}\} \sim q(\phi^{\tau}|\D^{\tau}; \psi)$
\State Initialize sub-learner with $\theta^{\tau}_0 = \theta * z^{\tau}$
\For{step in $1, ..., \mathsf{K}$}
\State Sample batch data from $\D^{\tau}_s$
\State Update parameters for task $\tau$ using $\theta^{\tau}_k = \theta^{\tau}_{k-1} - \gamma^{\tau} \circ \alpha \circ \sum_{c=1}^2 \omega_c^{\tau} \nabla_{\theta^{\tau}_{k-1}} \L(\theta^{\tau}_{k-1}, \D_s^{\tau})$
\EndFor
\State Sample batch data from $\D^{\tau}_t$
\State Evaluate $\L (\theta^{\tau}_{\mathsf{K}}, \D^{\tau}_t)$
\EndFor
\State Update meta-learner $f_{\theta}$ with $\theta = \theta - \beta \nabla_{\theta} \sum_{\tau=1}^{\mathcal{T}} \L (\theta^{\tau}, \D^{\tau}_t)$
\EndWhile
\State \textbf{Meta testing:} $Y^{\tau} \leftarrow f_{\theta}(X^{\tau}_{\mathsf{test}}, \mathcal{S}^{\tau}_{\mathsf{test}})$
\end{algorithmic}
\label{alg:few_shot_st}
\end{algorithm*}

\section{Task-Adaptive Style Transfer}
We discuss formulation of multi-pair text-style transfer problem using the TAML framework. An overview of our method is shown in Figure \ref{fig:showcase}. We assume training data in each task could either be parallel (task 1 and 4) or non-parallel (task 2 and 3). The number of training samples in task $i$ is represented by $N_i$, which is not necessarily equal for each task. In addition, the class distribution in non-parallel training data is heavily skewed. 

We now formulate our problem as follows. Given a distribution of similar tasks $p(\tau)$, each task represents performing text style transfer on a certain dataset $\D^{\tau}$. Define a generic loss function $\L$ and shared parameters $\theta$ within tasks, the goal is to jointly learn a task-agnostic model $f_{\theta}: (X^{\tau}, \mathcal{S}^{\tau})\mapsto Y^{\tau}$, where for each $\tau$, $\mathcal{S}^{\tau}$ is the corresponding set of style labels of original text $X^{\tau}$, and $Y^{\tau}$ is the resulting style transformed text. Ideally, $Y^{\tau}$ should be consistent with $\overline{X}^{\tau}$, the corresponding input text sentence in another style domain which may or may not be available in model training. In fine-tuning with a new task, the parameters are initialized accounting for the imperfect nature of the given dataset. Similar to the standard meta-learning approach, the training data of task $\tau$ is divided into a support set $\D^{\tau}_s$ and a query set $\D^{\tau}_t$, where $\D^{\tau}_s$ is used to update each sub-task and $\D^{\tau}_t$ is used to evaluate the loss, and later used for meta-learner updates. A detailed description can be found in Algorithm \ref{alg:few_shot_st}.

\section{Experiments}

We conduct experiments on multiple style-transfer datasets: Shakespeare \citep{xu2012paraphrasing}, Yelp reviews \citep{shen2017style} and an internal dataset from a company that contains formal/informal text sentences. Performing style transfer on each of the above dataset defines a unique task. The Shakespeare dataset contains 21k parallel sentences, which includes original text style and Shakespeare's style. The maximum length of the sentences is 20. The Yelp dataset contains around 252k sentences of positive and negative restaurant reviews, where we use a maximum length of 15 to conduct the experiment. We evaluate our method using state-of-the-art transformers including BERT, GPT-2, T5, and VAE \citep{john2018disentangled} designed for style transfer by learning disentangled representations. Our baseline method includes regular model training without distinguishing the difference between tasks, and the MAML method in Eq (\ref{eq:maml}) to fine-tune the style transfer models on multiple distinct tasks, which has also been proposed by \citet{chen2020st}. We then employ Algorithm \ref{alg:few_shot_st} to adaptively fine-tune the style transfer models for each task.

\begin{figure}[t]
    \centering
    \includegraphics[width=\linewidth]{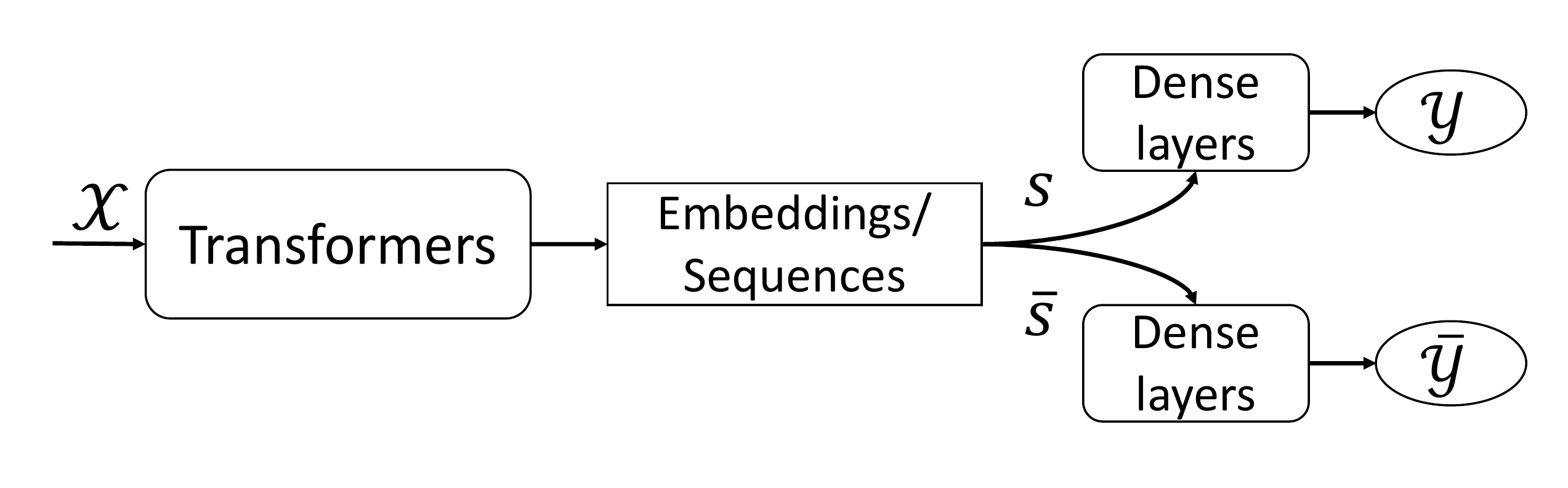}
    \caption{Two-head architecture}
    \label{fig:struc}
\end{figure}

 
The unbalanced training data is created by sampling from each class at different rate (75\% positive class, 25\% negative class). We use the pretrained transformers in Huggingface library \citep{wolf-etal-2020-transformers} as our initial style transfer models. Specifically, we build a two-head model (Figure \ref{fig:struc}) on top of the decoders where each head is composed of multiple dense layers. We do not perform end-to-end training for the entire transformer but only train the two-head model. The model input is the sentence and style pairs $(X^{\tau}, \mathcal{S}^{\tau})$ while the forward propagation of transformer's output to each model head is dependent on the style labels. The resulting output sentences are style dependent, and one can perform text-style transfer by flipping the style labels during the inference phase. Similarly, we use both baseline and TAML to train VAE and obtain disentangled style and content representations, and replace the style embedding during the inference stage to get style transferred sentences. Note that we focus on improving the fine-tuning part of text style transfer models, while we do not modify the model structure themselves. In terms of content preservation, the objective function of the VAE model proposed by \citep{john2018disentangled} contains a content-oriented loss, while for other transformer-based models, we designed the loss $\L$ to be the cross entropy loss between $Y^{\tau}$ and $\overline{X}^{\tau}$, or between $Y^{\tau}$ and $X^{\tau}$ in non-parallel situations.

For BERT, GPT-2, and T5, we use the built-in vocabulary within the transformers library. Adam optimizer is used with learning of $5\times10^{-4}$ to train the model. The batch size is set to 16 and the model is trained for 100 epochs. We build the two-head model by using 6 fully connected layers with hidden size of 256 and ReLU activation function. The parameters are chosen empirically with the best performance. For VAE approach, we use the same parameter settings as reported in \citep{john2018disentangled}. As for $\mathrm{NN_1}$ in inference network, we used two consecutive blocks of $3 \times 3$ convolution layer followed by $2 \times 2$ max pooling layer, the output is then fed into one fully connected layer for statistics pooling. We then use two fully connected layers for $\mathrm{NN_2}$. All the activation functions are ReLU.

\begin{table}[t]
\centering
\renewcommand\arraystretch{1.3}
\scalebox{0.66}{
\begin{tabular}{c|ccc|ccc}
\hlinewd{1.5pt}
\multirow{2}{*}{ Model } & \multicolumn{3}{c}{ Shakespeare } & \multicolumn{3}{c}{ Yelp } \\
\hhline{|~|---|---|}
& $\mathrm{BLEU}^{\uparrow}$ & $\mathrm{PPL}^{\downarrow}$ & $\mathrm{ACC}^{\uparrow}$ & $\mathrm{BLEU}^{\uparrow}$ & $\mathrm{PPL}^{\downarrow}$ & $\mathrm{ACC}^{\uparrow}$ \\
\hline
BERT & 12.04 & 26.43 & 78.77 & 9.56 & 15.31 & 74.68 \\
GPT-2 & 2.83 & 38.47 & 74.45 & 4.81 & 45.49 & 76.67 \\
T5 & 3.65 & 59.39 & 82.58 & 5.22 & 39.41 & 75.14 \\
VAE & 14.36 & 22.29 & 81.92 & 10.81 & 10.65 & 77.27 \\
\hhline{|=|===|===|}
MAML-BERT & 16.31 & 21.09 & 79.34 & 10.87 & 15.02 & 74.98 \\
MAML-GPT-2 & 7.01 & 36.94 & 75.25 & 5.04 & 41.76 & 77.06 \\
MAML-T5 & 4.77 & 50.44 & 83.02 & 6.46 & 33.72 & 75.86 \\
MAML-VAE & 15.52 & 21.45 & 81.96 & 11.74 & 11.04 & 77.24 \\
\hhline{|=|===|===|}
TAML-BERT & 17.56 & \textbf{19.36} & 79.34 & 11.02 & 16.82 & 75.22 \\
TAML-GPT-2 & 7.42 & 36.67 & 76.02 & 5.63 & 37.66 & 77.18 \\
TAML-T5 & 4.81 & 47.23 & \textbf{83.45} & 6.92 & 32.30 & 75.64 \\
TAML-VAE & \textbf{17.98} & 20.14 & 82.61 & \textbf{12.31} & \textbf{10.59} & \textbf{77.33} \\
\hlinewd{1.5pt}
\end{tabular}}
\caption{Evaluations of multiple text style transfer models on testing set of the listed data. TAML-based model training methods achieve better performance on multi-task text style transfer.}
\label{tab:res}
\end{table}

We evaluate competing methods on quality and accuracy of style transfer. The adopted metrics are common choices among recent works. 

\texttt{BLEU}: We use BLEU \citep{papineni2002bleu} score to evaluate the content preservation, the scores are calculated using ScareBLEU \citep{post-2018-call}. When parallel sentences are available, we compute the BLEU score between the style transferred sentences $Y^{\tau}$ and the ground truth sentences $\overline{X}^{\tau}$. Otherwise, we use the original sentences $X^{\tau}$ instead.

\texttt{PPL}: We implemented a bigram language model \citep{kneser1995improved} to quantitatively evaluate the fluency of a sentence. The language model is trained on the target-style domain, and we report the PPL of the generated sentences.

\texttt{Accuracy}: We also trained a TextCNN classifier \citep{rakhlin2016convolutional} simultaneously while training style transfer models. The trained classifier is then used to evaluate the classification accuracy on the generated sentences.

Table \ref{tab:res} shows our results for each method. By applying task-adaptive meta learning on each style-transfer model, the performance with respect to every metric is generally improved on the datasets we evaluated. We observe that the VAE method performs better in style transfer, as other models are not explicitly designed for this goal.


\section{Conclusion}
In this paper, we investigated meta-learning approaches applied to text-style transfer, for situations with multiple data sources. Given the distinct context and total amount of data, we propose a task-adaptive meta-learning approach to fine-tune style-transfer models. The proposed method introduces three balancing variables with probabilistic distributions, which can be encoded from training data. These balancing variables are then used to solve class and task imbalance problems. Empirically, we found that TAML improves the style-transfer performance on multiple models. In the future, we wish to explore generating style variations in more fine-grained levels (for $C > 2$) with the help of meta-learning.

\bibliographystyle{acl_natbib}
\bibliography{reference}
\end{document}